\documentclass[conference]{IEEEtran}

\usepackage{times}

\usepackage[numbers,sort]{natbib}
\usepackage{multicol,multirow}
\usepackage[bookmarks=true]{hyperref}
\usepackage{graphicx} 
\usepackage{subcaption} 
\usepackage[dvipsnames]{xcolor} 
\usepackage{amsmath,amssymb} 
\usepackage{stfloats} 
\usepackage{booktabs, makecell} 
\usepackage{mfirstuc,titlecaps} 
\usepackage{etoolbox}
\usepackage{algorithm,algpseudocode}

\title{Gain Tuning Is Not What You Need: Reward Gain Adaptation for Constrained Locomotion Learning}

\author{Arthicha Srisuchinnawong$^1$ and Poramate Manoonpong$^{1,2}$ \\ {\small $^1$Vidyasirimedhi Institute of Science and Technology, Rayong, Thailand.} \\ {\small $^2$The University of Southern Denmark, Odense, Denmark.}}

\begin{document}

\def \roger{ROGER}
\def \rogerf{Reward-Oriented Gains via Embodied Regulation}

\def \deg{$^\circ$}

\newcommand\pval[2]{\textit{(p-value #1, #2)}}

\newcommand\rd[1]{\textcolor{red}{#1}}
\newcommand\bu[1]{\textcolor{blue}{\textbf{#1}}}
\newcommand\cy[1]{\textcolor{blue}{\textbf{#1}}}

\def \main{primary}
\def \cthres{constraint threshold}

\newcommand\var[2]{\vart{#1}{#2}{t}}
\newcommand\tvar[2]{\text{\var{#1}{#2}}}
\newcommand\vart[3]{$#1_{#2#3}$}
\newcommand\tvart[3]{\text{\vart{#1}{#2}{#3}}}
\newcommand\e[1]{10$^\text{-#1}$}


\def \simbonevideo{\href{https://youtu.be/cZ5qOw0i_T4}{\url{https://youtu.be/cZ5qOw0i_T4}}}
\def \realbonevideo{\href{https://youtu.be/F1olq7W6J9g}{\url{https://youtu.be/F1olq7W6J9g}}}
\def \realunevenvideo{\href{https://youtu.be/Cqu7vLT_Piw?si=jtzJCpRubbFHx06w}{\url{https://youtu.be/Cqu7vLT_Piw?si=jtzJCpRubbFHx06w}}}
\def \hoppervideo{\href{https://youtu.be/SpL4awVgDZM}{\url{https://youtu.be/SpL4awVgDZM}}}
\def \github{\href{https://github.com/Arthicha/ROGER_ROGER_public}{\url{https://github.com/Arthicha/ROGER_ROGER_public}}}

\newcommand\revise[1]{#1}

\makeatletter
\apptocmd{\@maketitle}{\bigskip \centering
		\includegraphics[width=0.95\textwidth]{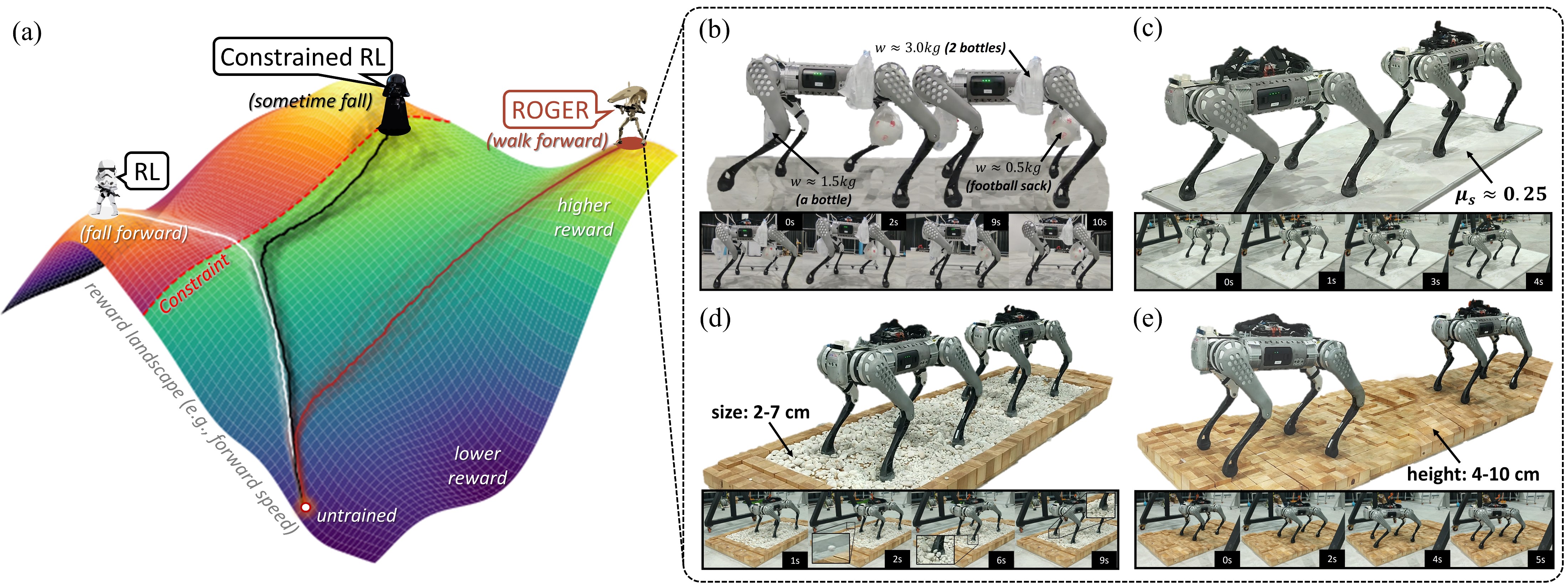}
		\captionof{figure}{(a) Parameter trajectories from (white) RL, (black) constrained RL, and (brown) \roger{} on a simulated reward landscape with their (transparent) explorations. Brighter regions indicate higher rewards, while darker regions indicate lower rewards. The red areas highlight violations with a red dashed line indicating the \cthres. RL and constrained RL consistently violate constraints, possibly during exploration, while \roger{} effectively avoids violations. \revise{Physical quadruped locomotion learning from scratch (b) with dynamic load, (c) on slippery terrain, (d) on a loose gravel field, and (e) on a random step field. A video of this experiment is available at \realunevenvideo.}} \label{fig:intro}
}{}{}
\makeatother

\maketitle

\begin{abstract}
	
Existing robot locomotion learning techniques rely heavily on the offline selection of proper reward weighting gains and cannot guarantee constraint satisfaction (i.e., constraint violation) during training. Thus, this work aims to address both issues by proposing \rogerf{} (\roger), which adapts reward-weighting gains online based on penalties received throughout the embodied interaction process. The ratio between the positive reward (\main{} reward) and negative reward (penalty) gains is automatically reduced as the learning approaches the \cthres s to avoid violation. Conversely, the ratio is increased when learning is in safe states to prioritize performance. With a 60-kg quadruped robot, \roger{} achieved near-zero constraint violation throughout multiple learning trials. It also achieved up to 50\% more \main{} reward than the equivalent state-of-the-art techniques. In MuJoCo continuous locomotion benchmarks, including a single-leg hopper, \roger{} exhibited comparable or up to 100\% higher performance and 60\% less torque usage and orientation deviation compared to those trained with the default reward function. Finally, real-world locomotion learning of a physical quadruped robot was achieved from scratch within one hour without any falls. Therefore, this work contributes to constraint-satisfying real-world continual robot locomotion learning and simplifies reward weighting gain tuning, potentially facilitating the development of physical robots and those that learn in the real world.
\end{abstract}

\setcounter{figure}{1}


\section{Introduction}

Robot locomotion is a challenging task involving \revise{the} embodied interaction between the robot and the environment \cite{embodiedai,howbodyshapethewaywethink}. To properly exploit this embodied interaction, reinforcement learning (RL) has been employed as a promising framework, enabling robots to discover effective control policies on their own \cite{rlbook}. As a result, this technique has demonstrated remarkable success \cite{anymalparkour,rapidlocomotion,notonlyrewardbutalsoconstraints}. Nevertheless, traditional RL approaches are influenced by random exploration and are prone to constraint violation. Desired characteristics, such as being stable without falling and staying within physical limits, cannot be guaranteed either during \cite{walkinthepark} or after \cite{notonlyrewardbutalsoconstraints,iros_ethcompare} the learning process (white path in Figure~\ref{fig:intro}), leading to instability, danger, and degraded performance \cite{saferl_review}.

To deal with this challenge, constrained RL emerges as a technique to enforce the constraints \revise{through} mathematical formulations describing the desired characteristics of robot behaviors \cite{saferl_review}. This technique can be divided into model-based and model-free approaches. The model-based approach relies on accurate models for learning, making model-free more practical. Without an environment model, model-free constrained RL represents undesired behaviors using penalty terms (\var{R}{i}), resulting in the following reward function (\var{R}{}):
\begin{equation}
\tvar{R}{} = \tvar{\lambda}{0}\tvar{R}{0} - \sum \tvar{\lambda}{i}\tvar{R}{i}, \label{eq:totalreward}
\end{equation} where \var{R}{}, \var{R}{0}, and \var{R}{i} denote the total reward, \main{} reward (i.e., \revise{the} weighted summation of all positive reward terms), and $i^{th}$ semi-positive definite constraint penalty terms at time $t$, while \var{\lambda}{0} and \var{\lambda}{i} denote the corresponding weighting gains. This approach can be categorized further into two groups: fixed-weighting, where the gains are pre-tuned and fixed; and adaptive weighting, where the gains are adapted or learned.

\subsection{Fixed-Weighting Constrained RL}

In fixed-weighting constrained RL, \var{\lambda}{0} and \var{\lambda}{i} are carefully tuned offline, either empirically or by hyperparameter searching, and kept fixed throughout the learning. Most works employ fixed weighting gains with error-based penalties, e.g., a squared error \revise{relative to the} desired states or a highly negative penalty given at undesired states to enforce constraint satisfaction and the desired results. Others model the penalty terms as control barrier functions (CBFs) \cite{notonlyrewardbutalsoconstraints,cbf}, providing high penalties near safety boundaries only to obtain improved performance. These techniques could include as many as 16 terms with 11 distinct properly selected values \cite{anymalparkour}, where adjusting a single hyperparameter may require four additional training repetitions \cite{hyperandhowtotunethem}. Choosing improper values may result in unnatural motion, undesired gaits, or poor performance \cite{notonlyrewardbutalsoconstraints}, making the selection process crucial and time-consuming.

\subsection{Adaptive Weighting Constrained RL}

To tackle the issue of fixed-weighting constrained RL, in adaptive weighting constrained RL, \var{\lambda}{0} and \var{\lambda}{i} are continuously adjusted throughout the learning. \revise{Adaptive weighting constrained RL can be further divided into two categories: (1) primal-dual approaches \cite{dd,cpo,trustqp_gofeasible,ipo}, which optimize the network parameters in the primal update while numerically adjusting the penalty regularization gains in the dual update, and (2) primal-only approaches \cite{crpo,distorted_distribution}, which explicitly compute the penalty regularization gains rather than using iterative optimization.} 

One early technique, primal-dual optimization (PDO) \cite{dd}, maximizes the total reward in the primal update while simultaneously adjusting the penalty gains in the dual update, as summarized in Appendix. This method provides a foundation for later works like constrained policy optimization (CPO) \cite{cpo}, which resets the penalty gains and incorporates trust regions for each dual update to achieve higher performance with greater computational complexity. Although a recent quadruped robot trained with CPO managed to progressively reduce its orientation deviation during learning, its exploration trials still experienced constraint violations \cite{trustqp_gofeasible}. Building upon CPO, inertia-point policy optimization (IPO) \cite{ipo} replaces penalty terms with logarithmic CBF; however, a quadruped robot trained with IPO experienced around 1,300 constraint violations per 1,000 episodes, according to \cite{iros_ethcompare}. 

One possible reason for constraint violation during learning is the delay in adjusting the regularization gains \revise{in primal-dual approaches} \cite{crpo}. In other words, \revise{multiple} updates are required to increase a penalty gain from nearly zero to a high proper value to ensure constraint satisfaction. To investigate this, constraint-rectified policy optimization (CRPO) \cite{crpo}, \revise{a primal-only approach,} introduces a reward-switching mechanism, which performs policy updates solely with the constraint penalties after a certain tolerance is exceeded, as summarized in Appendix
. \revise{Although \cite{iros_ethcompare} reported that} CRPO outperformed previous techniques, like CPO and IPO, a quadruped robot trained with CRPO still experienced around 960 violations per 1,000 episodes; besides, its effectiveness relied on the switching threshold, i.e., tolerance parameter. Following that, according to \cite{distorted_distribution}, a variant of CRPO incorporates a distortion-based risk measure to modify reward distribution, highlighting high-penalty actions; however, a quadruped robot trained with this technique still experienced around 300 constraint violations per 1,000 episodes. Despite simplifying much of the tuning process, constraint violations persist to some extent (black path in Figure~\ref{fig:intro}). 

\revise{Recent techniques proposed in early 2025 are based on these approaches. For instance, QRSAC-Lagrangian \cite{QRSAC} performs dual updates as in PDO with Adam optimizer to handle shift in value functions and improve stability, Constraint-Rectified Multi-Objective Policy Optimization (CR-MOPO) \cite{CRMOPO} modifies CRPO for multiple constraint objectives along with incorporating a conflict-averse technique optimized with natural policy to deal with conflicting objectives, while \cite{SAPIECBF} uses neural network-based CBFs trained to adjust CBFs for specific conditions.}

\revise{Recent techniques proposed in early 2025 build on these approaches. For instance, QRSAC-Lagrangian \cite{QRSAC} performs a dual update similar to PDO, using the Adam optimizer to handle shifts in value functions and improve stability. Constraint-Rectified Multi-Objective Policy Optimization (CR-MOPO) \cite{CRMOPO} extends CRPO to multiple constraint objectives and incorporates a conflict-averse technique optimized with a natural policy gradient to handle conflicting objectives. Meanwhile, \cite{SAPIECBF} employs neural network-based CBFs that are trained to adapt to specific conditions.}

\subsection{Current Stage of Real-World Locomotion Learning}

Although several constrained RL techniques have been proposed, the guarantee of near-zero constraint violations, especially during learning, has \revise{yet to be} achieved. Only two robots have demonstrated locomotion learning in the real world by using simple reward functions with some workarounds \cite{smeagol,walkinthepark}. In one work, a physical hexapod robot achieved locomotion learning within 20 minutes using a single-term speed reward function without falling, thanks to the stability provided by leg redundancy \cite{smeagol}. In another work, a physical quadruped robot achieved this under a similar time scale using a reward function combining speed and orientation plus a fall recovery policy for resetting the robot \cite{walkinthepark}. Therefore, when foot contact redundancy cannot be obtained or the consequence of falling is destructive, safety constraints defined using certain robot state variables, such as orientation, must be predefined, included in the reward function, and satisfied at all times. 

\begin{figure}[!h]
	\centering
	\includegraphics[width=\linewidth]{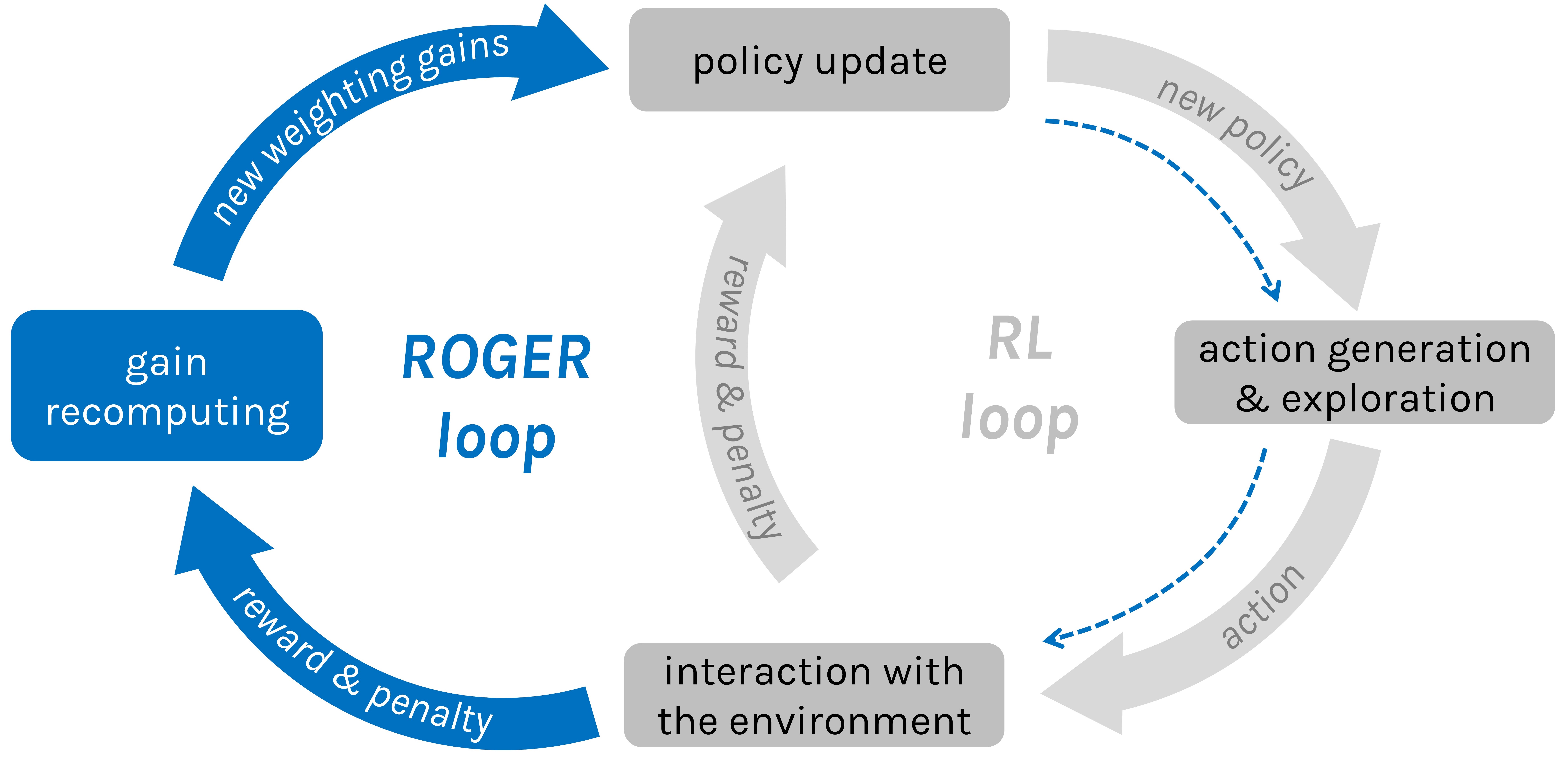}
	\caption{Illustration of how embodied interaction between the robot and \revise{the} environment can be used to train \revise{a} control policy. The traditional RL loop is shown in gray, and the additional \roger{} loop \revise{is shown} in blue.}
	\label{fig:rogerconcept}
\end{figure}

To this end, this work hypothesizes that robot-environment interaction can be leveraged for both policy updates (RL loop) and dynamic reward gain adjustments (online weighting gain adaptation loop), as illustrated in Figure~\ref{fig:rogerconcept}, without introducing additional \revise{hyperparameters that require tuning}. Building on this idea, this work proposes \rogerf{} (\roger), dynamically balancing reward weightings based on proximity to the \cthres s. These weightings are continuously refined through embodied interaction, where insufficient penalty gains drive the robot toward unsafe states, prompting stronger penalties and reducing the \main{} gain to prevent violations. In summary, \roger{} has the following key advantages:
\begin{enumerate}
	\item Ensuring constraint satisfaction throughout learning (brown path in Figure~\ref{fig:intro}).
	\item Elimination of extensive gain tuning by utilizing intuitive hyperparameters, e.g., \cthres s.
\end{enumerate}

This paper is organized as follows: Section~\ref{sec:method} describes \roger{} in more detail; Section~\ref{sec:exp} presents the experimental results, including real-world RL of a quadruped robot; Section~\ref{sec:conclusion} discusses the results and future potential; and finally, Section~\ref{sec:limit} discusses the limitations.

\section{Reward-Oriented Gains via Embodied Regulation (ROGER)}
\label{sec:method}

The core idea of \rogerf{} (\roger) is to ensure safe operation while maximizing performance by dynamically adjusting the reward structure through real-time interaction with the environment, \revise{thereby} eliminating the need for additional non-intuitive hyperparameters. When the robot is far from the \cthres s, \roger{} uses the high \main{} reward weighting gain with low penalty weighting gains to encourage task-specific optimization. As the robot approaches the \cthres s, \roger{} automatically increases the penalty weighting gains and reduces the \main{} reward weighting gain, discouraging unsafe actions and reducing the penalties. This process continues iteratively: when penalties reach a critical point, they steer the robot back into safe regions, after which the penalty weighting gains decrease, allowing for improved performance.

\begin{figure}[!h]
	\centering
	\begin{subfigure}[b]{0.45\linewidth}
		\centering
		\includegraphics[width=\linewidth]{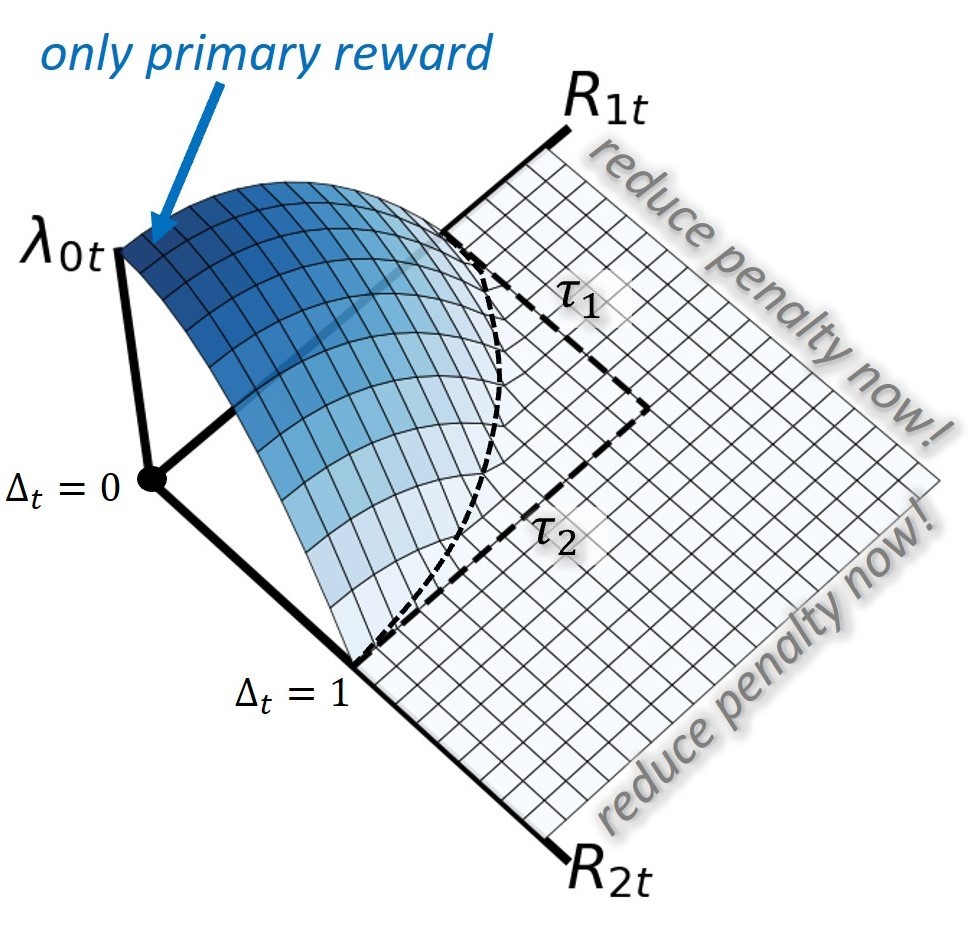}
		\caption{}
		\label{fig:adaptsurface_rwd}
	\end{subfigure}
	\hfill
	\begin{subfigure}[b]{0.45\linewidth}
		\centering
		\includegraphics[width=\linewidth]{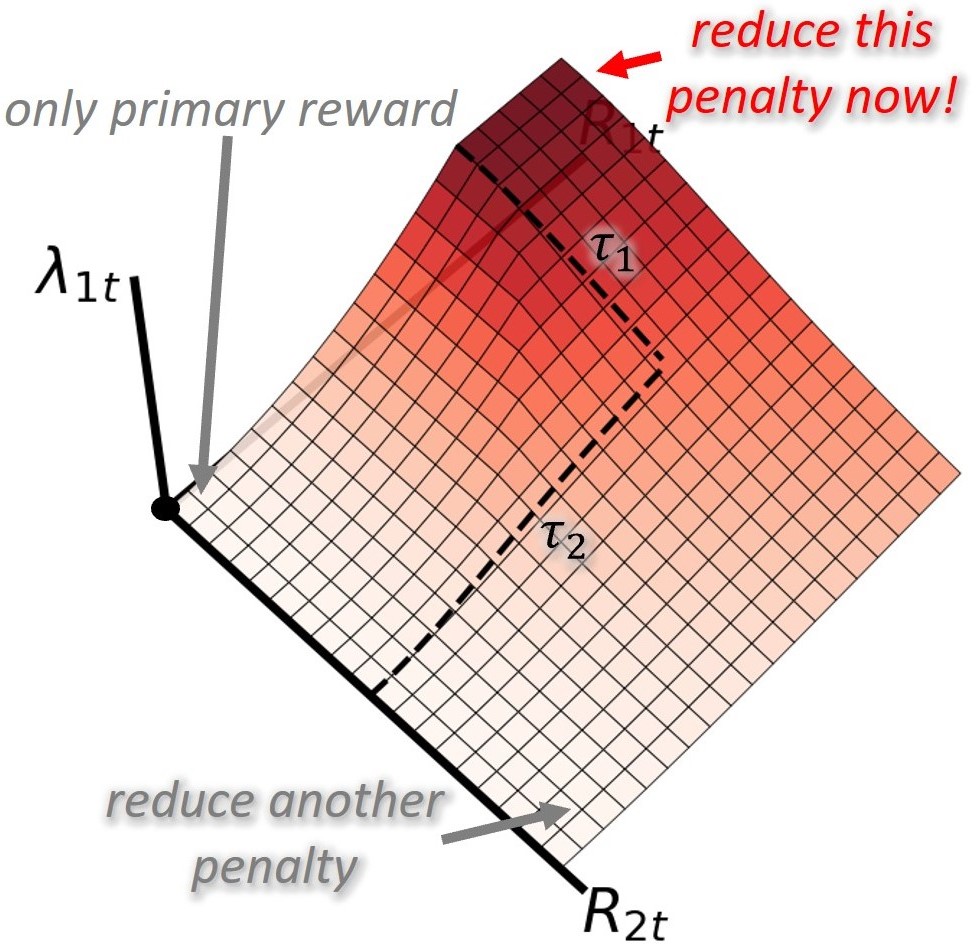}
		\caption{}
		\label{fig:adatsurface_penal}
	\end{subfigure}
	\caption{
		This adaptation strategy is also proven to be partially stable in key conditions (i.e., near the \cthres s and convergence), while the expected \main{} reward is guaranteed to increase, as detailed in Appendix. As a result, if an optimal solution exists far from the \cthres s, such as body orientation in robot locomotion, \roger{} will converge to that solution; otherwise, it chooses a safe alternative.
	}
	\label{fig:adaptsurface}
\end{figure}

To achieve this, \roger{} adopts a multiple-reward-channel setup \cite{multirewardchannel}, where rewards and penalties are stored independently before being combined into total advantage scores for policy updates. This setup enables \roger{} to dynamically adjust the reward and penalty weighting gains (Equation~\ref{eq:totalreward}) across different learning episodes and timesteps. At each timestep, the weighting gains are computed according to:
\begin{align}
\tvar{\lambda}{0} &= 1 - \Delta_t, \label{eq:roger0}\\
\tvar{\lambda}{i} &= \tvar{r}{\lambda_i} \Delta_t ,  \label{eq:roger1}
\end{align} where $\Delta_t$ denotes the clipped summation of constraint penalties, preventing $\tvar{\lambda}{0} < 0$ and ensuring that $\tvar{\lambda}{0} + \Sigma \tvar{\lambda}{i} = 1$, and \var{r}{\lambda_i} denotes the constraint contribution ratio of the $i^{th}$ constraint penalty term. $\Delta_t$ and \var{r}{\lambda_i} are computed from the ratio of the estimated constraint penalties (\var{\tilde{R}}{i}) to the \cthres s ($\tau_i$) as:
\begin{equation}
\Delta_t = \operatorname{min} \left\{ \text{$\sum_{j}$} (\tvar{\tilde{R}}{j}/\tau_j)^2 ,1.0 \right\},  \label{eq:deltasum}
\end{equation}
\begin{equation}
\tvar{r}{\lambda_i} = \frac{(\tvar{\tilde{R}}{i}/\tau_i)^2}{\sum_{j} (\tvar{\tilde{R}}{j}/\tau_j)^2}, \label{eq:ratio}
\end{equation} where $\tau_i$ and $\tau_j$ are selected intuitively based on the physical properties of the system, e.g., maximum hardware limits, together forming a safe region into which the robot should be bounded, as shown by the dashed lines in Figure~\ref{fig:adaptsurface}, while \var{\tilde{R}}{i} is \revise{a statistical estimate of the $i^\text{th}$ penalty at time $t$: $\tvar{\tilde{R}}{i} = \text{average}[\tvar{{R}}{i}] - k_\sigma \text{std}[\tvar{\sigma}{i}]$. A preliminary result when varying $k_\sigma$ is presented in the Appendix.}

This adaptation strategy is also proven to be partially stable in key conditions (i.e., near the \cthres s and convergence), while the expected \main{} reward is guaranteed to increase, as detailed in Appendix \cite{lyapunov}. As a result, if an optimal solution exists far from the \cthres s, such as body orientation in robot locomotion, \roger{} will converge to that solution; otherwise, it chooses a safe alternative.

Intuitively, the robot initially maximizes the \main{} reward since all penalty terms and their gains are zero, i.e., $\Delta_t = 0$ and $\tvar{\lambda}{0} = 1$, as illustrated in Figure~\ref{fig:adaptsurface_rwd}. As the robot approaches a \cthres, the \var{R}{i} terms increase, \revise{causing} $\Delta_t$ to grow, reducing \var{\lambda}{0} and increasing \tvar{\lambda}{i}, \revise{thereby distributing} more and more attention to the penalty terms. This setup prevents the robot from approaching the \cthres s. However, when near a \cthres, i.e., some \tvar{R}{i} $ \approx \tau_i$, $\Delta_t$ saturates at 1.0, making $\tvar{\lambda}{0} \approx 0$ and $\tvar{R}{} \approx -\tvar{R}{i}$. This setup neglects the \main{} reward, immediately enforcing the reduction of the penalty terms and \revise{ensuring} constraint satisfaction.

Simultaneously, the penalty weighting gains are distributed according to the ratio of \var{R}{i}, as illustrated in Figure~\ref{fig:adatsurface_penal}. As the first penalty term (\var{R}{1}) increases, it receives more weight and higher optimization priority. However, as other terms, e.g., \var{R}{2}, rise, the weighting gains are redistributed, lowering the priority of the first penalty term to increase those of the others. Thus, this setup automatically balances the contributions between different penalty terms online during locomotion learning. \revise{In total, \roger{}\footnote{\revise{ Code is available at \github.}} takes an additional computation time of 0.46$\pm$0.09 ms on an Intel i7 CPU with an Nvidia GTX1050 GPU, or approximately 0.03\% of data collection/exploration time.}

\section{Experiments and Results} 
\label{sec:exp}
Two experiments were conducted to investigate and evaluate \roger. The first compared \roger{} with state-of-the-art techniques on a 60-kg quadruped robot in simulation, followed by the demonstration of real-world locomotion learning. The second further tested \roger{} with a MuJoCo hopper, which violates the assumptions presented in Appendix: zero-penalty optimality and gentle system and transition dynamics. Under this condition, the reward weightings adapted using \roger{} were compared with the default reward weightings provided by OpenAI Gymnasium \revise{and CRPO}. The results of other locomotion learning tasks are provided in Appendix. 

\subsection{Quadruped Locomotion Learning}

\begin{figure}[!h]
	\centering
	\includegraphics[width=\linewidth]{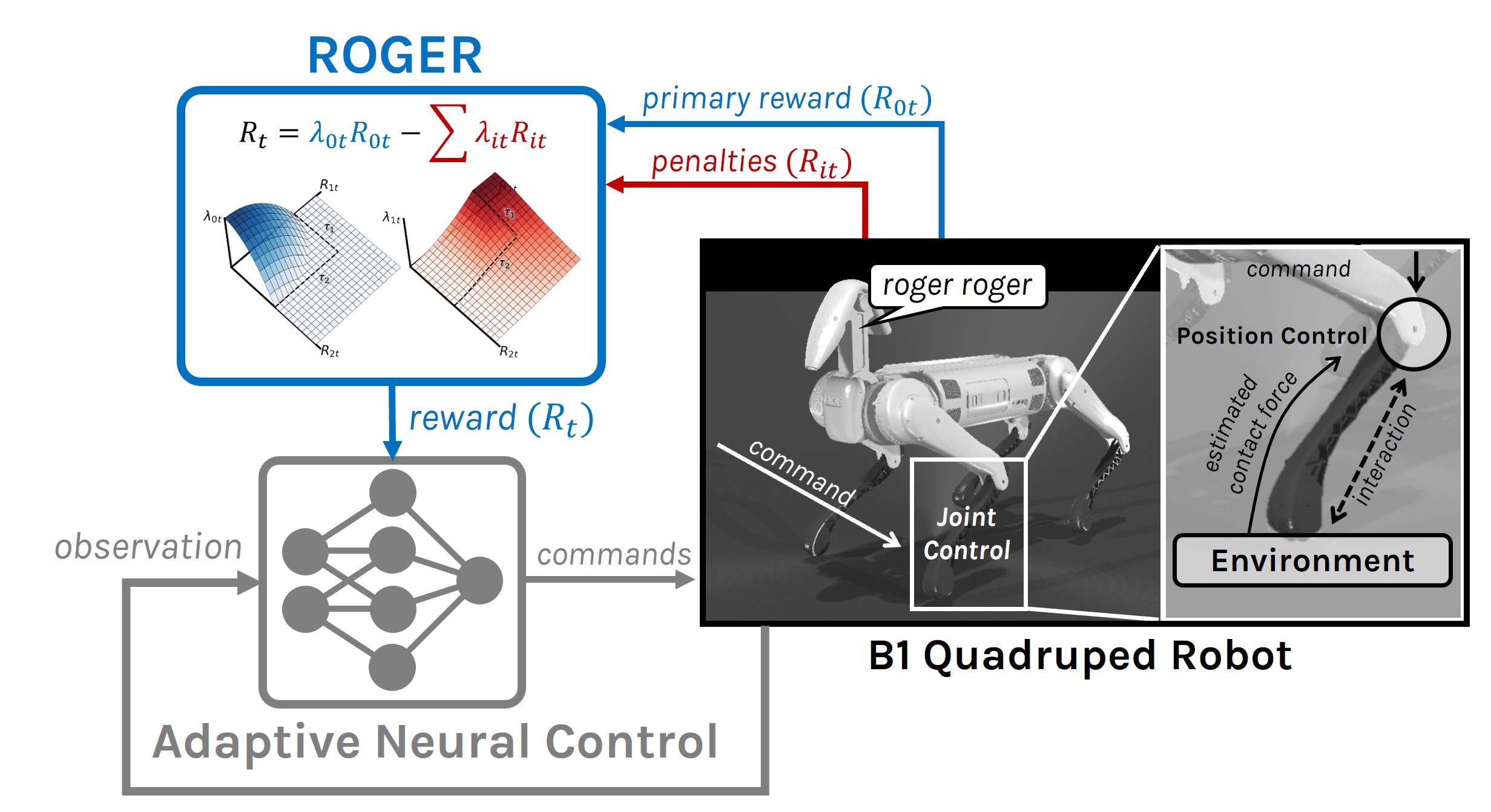}
	\caption{Locomotion learning framework for the Unitree B1 quadruped with \roger. An adaptive neural control produces joint position targets \revise{that are} used by low-level controllers. After execution, the robot receives rewards and penalties, which are then combined using \roger{} and \revise{subsequently} used to train the neural control.}
	\label{fig:quaddiagram}
\end{figure}

Unlike previous studies that used small, less sensitive quadruped robots, the first experiment performs a comparison on a 60-kg Unitree B1 (Figure~\ref{fig:quaddiagram}), a platform more prone to instability and falling if improper policies are learned. To ensure safe and efficient locomotion learning, a state-of-the-art adaptive neural control framework, called Sequential Motion Executor-Adaptive Gradient-weighting Online Learning (SME-AGOL) \cite{smeagol}, was adopted, as shown in Figure~\ref{fig:quaddiagram}. This control framework \revise{has} previously demonstrated real-world locomotion learning on a physical hexapod robot. \revise{Moreover}, it supports stable and smooth locomotion by performing exploration in parameter space, e.g., avoiding noisy exploration as in PPO, and providing interpretability, e.g., mapping weights directly to a series of corresponding robot trajectories/configurations. 

\begin{figure*}[!t]
	\begin{subfigure}[t]{0.28\linewidth}
		\centering
		\includegraphics[width=\linewidth]{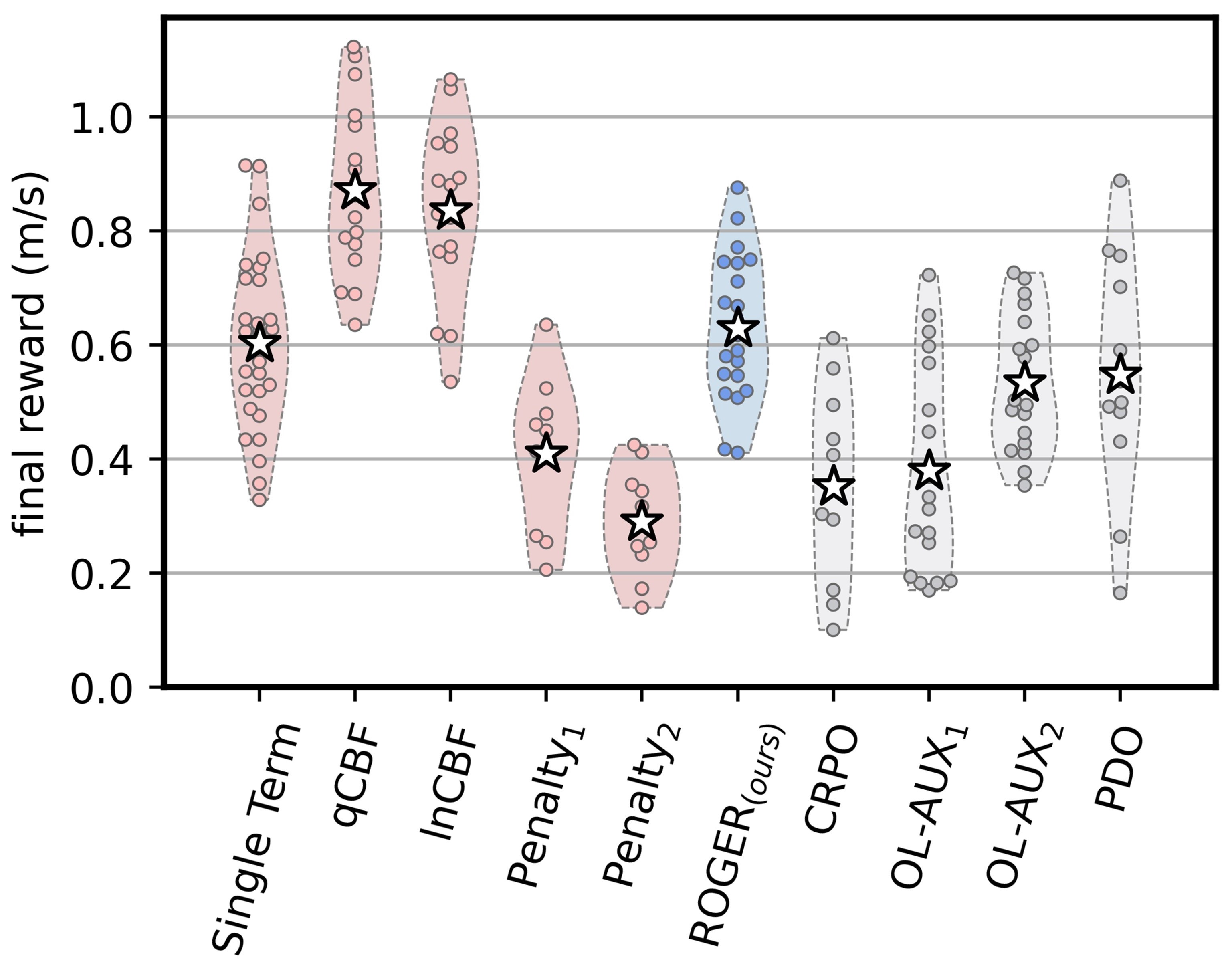}
		\caption{}
		\label{fig:simquadreward}
	\end{subfigure}
	\hfill
	\begin{subfigure}[t]{0.31\linewidth}
		\centering
		\includegraphics[width=\linewidth]{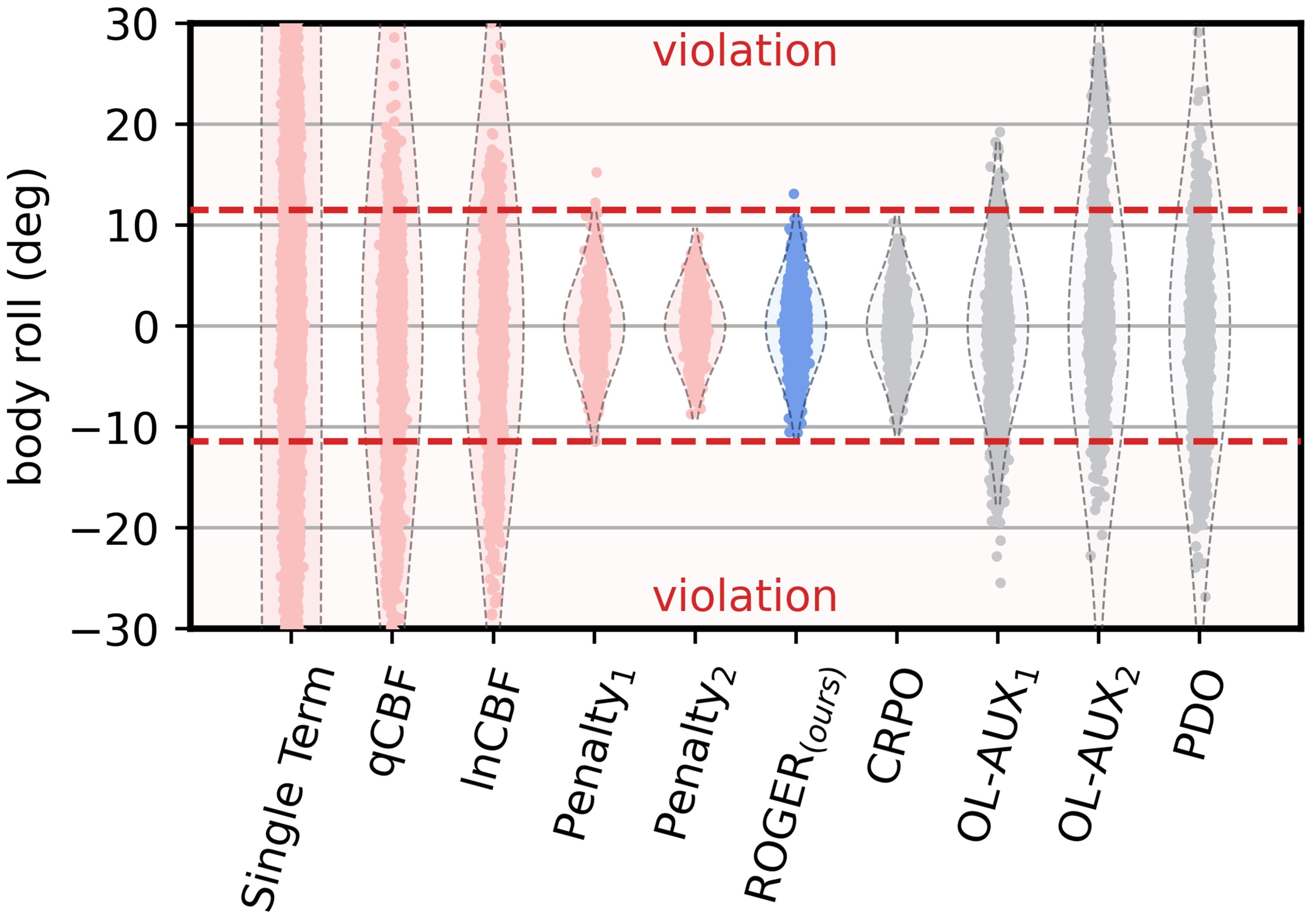}
		\caption{}
		\label{fig:simquadroll}
	\end{subfigure}
	\hfill
	\begin{subfigure}[t]{0.31\linewidth}
		\centering
		\includegraphics[width=\linewidth]{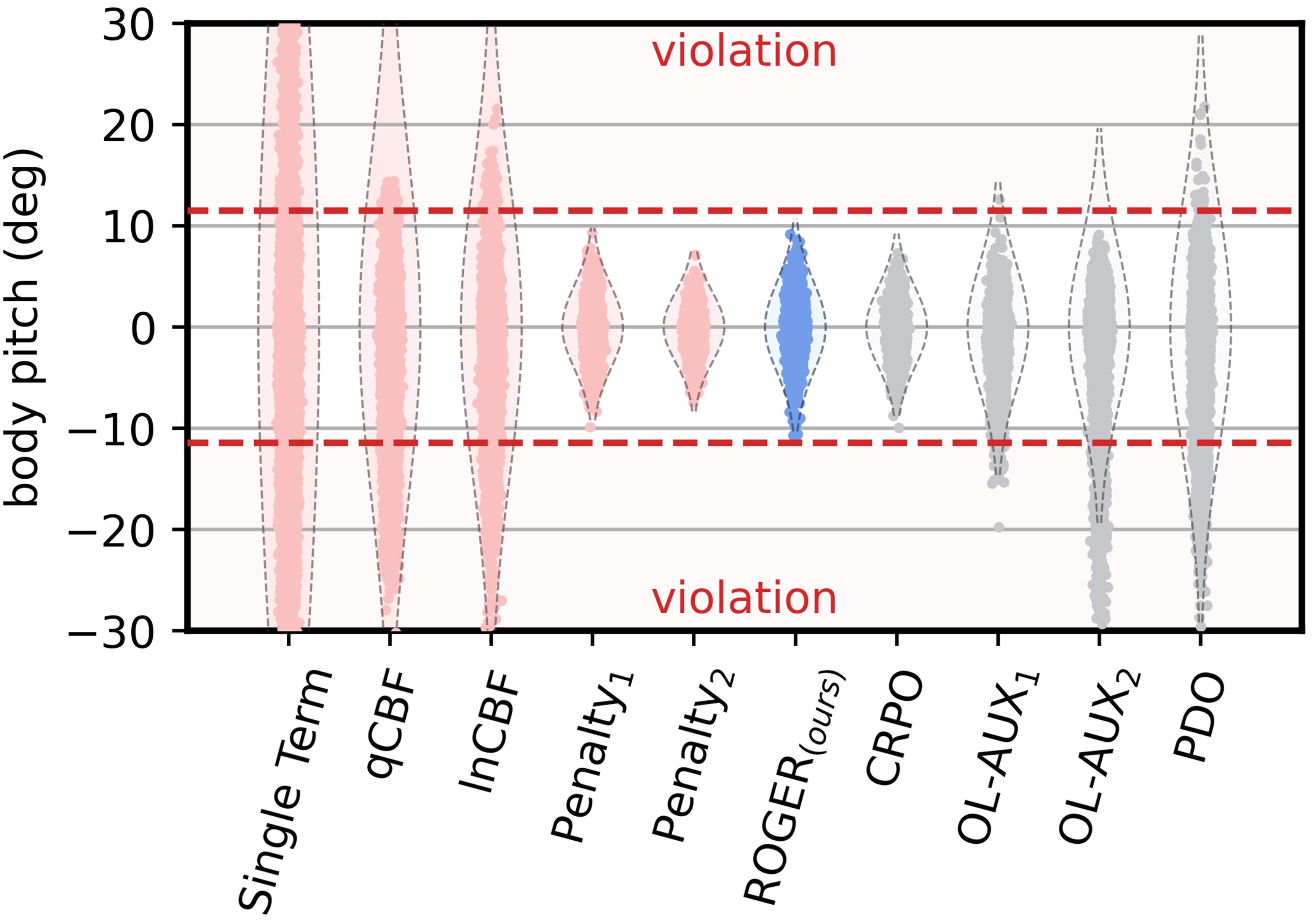}
		\caption{}
		\label{fig:simquadpitch}
	\end{subfigure}
	\caption{(a) Final \main{} reward values obtained from the last training episode and (b-c) roll and pitch angles recorded throughout the locomotion learning of the simulated Unitree B1 quadruped robot. The robot was trained using six techniques: two fixed-weighting \revise{techniques} in red (fixed-gain penalty and fixed-gain CBF), three adaptive weighting \revise{techniques} in gray (PDO, CRPO, and OL-AUX), and \roger{} in blue. All conditions are presented along with their kernel density estimation. In (b-c), red dashed lines indicate \cthres s at $\pm$ 0.2 rads, or approximately $\pm$ 10\deg; therefore, the data points exceeding these lines indicate violations. A video of this experiment is available at \simbonevideo.}
	\label{fig:simquadresult}
\end{figure*}

The SME network is a central pattern generator-based network. It consists of three layers: (1) a central pattern generator layer producing rhythmic activity, (2) a triangular basis layer forming locomotion bases, and (3) an output layer mapping the bases to eight joint position commands \revise{for} the quadruped \revise{robot}. It should be noted that \revise{the} hip abduction/adduction joints were fixed, reserving them for future turning control. The mapping weights were initialized \revise{to} zero before being optimized \revise{using} the AGOL learning rule as follows:
\begin{equation}
\Delta \theta = \eta_\theta \sum_{i=0}^{n} |\nabla_\theta \tilde{a}_i| \left(\frac{\tilde{\theta}_i - \theta}{\sigma_\theta^2}\right) \left(\frac{ G_i - \bar{G}_i }{\sigma_{G_i}}\right),
\end{equation} 
\begin{equation}
\Delta \sigma_\theta = \eta_\sigma \sum_{i=0}^{n} |\nabla_\theta \tilde{a}_i| \left(\frac{(\tilde{\theta}_i - \theta)^2 - \sigma_\theta^2}{\sigma_\theta^3}\right) \left(\frac{ G_i - \bar{G}_i }{\sigma_{G_i}}\right),
\end{equation}

\noindent where $\Delta \theta$ and $\Delta \sigma_\theta$ denote the updates to \revise{the} parameters ($\theta$) and \revise{the} exploration standard deviations ($\sigma_\theta$), respectively. $\eta_\theta$ and $\eta_\sigma$ denote the learning rates. $n$ denotes the length of the stored trajectories, i.e., 8 episodes $\times$ 70 timesteps ($\approx$ 3 gait cycles per episode). $\nabla_\theta \tilde{a}_i$ denotes the gradient of the explored action\revise{s}. $\tilde{\theta}$ denotes the explored parameter \revise{re-randomized} at the beginning of every episode. $\bar{G_i}$ and $\sigma_{G_i}$ denote the average and standard deviation of the return $G_i$, respectively, serving as the baseline and normalization gain\revise{s} for advantage estimation. Taking advantage of the rhythmic nature of locomotion, the return $G_i$ is computed from the average of the total reward performed over a horizon of 20 timesteps, or approximately a gait cycle. Here, \roger{} was applied as an add-on mechanism to the AGOL learning mechanism by adapting its reward weighting gains.

The total reward function in Equation~\ref{eq:totalreward} includes \revise{a} forward velocity reward ($\tvar{R}{0} = v_\text{fwd}$, encouraging forward movement) and orientation penalties for roll ($\tvar{R}{1} = |\alpha_t|$, penalizing sideways oscillation) and pitch ($\tvar{R}{2} = |\beta_t|$, penalizing longitudinal oscillation). The \cthres s, e.g., $\tau_j$ in Equations~\ref{eq:deltasum} and \ref{eq:ratio}, are selected as 0.2 rad, or approximately 10\deg. To include 99.9\% of \revise{the} exploration uncertainty in the process, \roger{} recomputes the reward weighting gains using \revise{$k_\sigma = 3$.} In other state-of-the-art methods, $\tvar{\tilde{R}}{i}$ is used as summarized in Appendix.

\subsubsection{Simulated Robot Experiment} The comparison between \roger{} and state-of-the-art constrained RL techniques was \revise{conducted} using a Mujoco-based simulation of a 60-kg Unitree B1 quadruped robot. The state-of-the-art techniques include two fixed-weighting \revise{techniques}: fixed-gain penalty (Penalty) and fixed-gain CBF (qCBF: quadratic CBF and lnCBF: logarithmic CBF); three adaptive-weighting \revise{techniques}: PDO, CRPO, and OL-AUX; and \roger. Note that OL-AUX \cite{ol_aux} is included here as an additional baseline because it dynamically adapts the auxiliary weighting gains (e.g., penalty), with the aim of exploiting the auxiliary terms to maximize the \main{} reward. The summarized details of all techniques are provided in Appendix. 

The key metrics for the evaluation are the final \main{} reward (i.e., robot forward speed), the constraint violation rate \revise{throughout the learning process obtained} from kernel density estimation, and the maximum roll and pitch deviations. For each technique, hyperparameters were selected through a grid search \cite{hyperandhowtotunethem}, with \revise{those yielding high rewards and low constraint violations} being selected for comparison. Some of the preliminary selection results are presented in Appendix. The robot was subjected to 500 training episodes under each condition, with over 10 repetitions per condition.

Figures~\ref{fig:simquadroll} and \ref{fig:simquadpitch} present that, in this case, the fixed gain penalty is the only fixed-weighting approach capable of \revise{satisfying the constraints} throughout learning. When properly selected, the fixed gain penalty \revise{resulted in} below 10\deg{} of orientation deviation with an estimated violation probability of $2\times10^{-12}$, or three violations in 25,000 timesteps. In contrast, training the quadruped robot solely with the \main{} reward did not satisfy any constraints: the robot \revise{experienced} over 30\deg{} orientation deviations with an estimated violation probability of 0.85, \revise{both of which were} significantly higher than those of the fixed-gain penalty \pval{$\ll$ 0.01}{two-proportion test}. Similarly, CBF-based approaches, whether logarithmic or quadratic, reduced the estimated violation probability to 0.60--0.16, with most of the deviation being below 20\deg; yet, the numbers were still significantly higher than those of the fixed gain penalty \pval{$\ll$ 0.01}{two-proportion test}.

 \begin{figure}[h]
	\centering
	\includegraphics[width=0.9\linewidth]{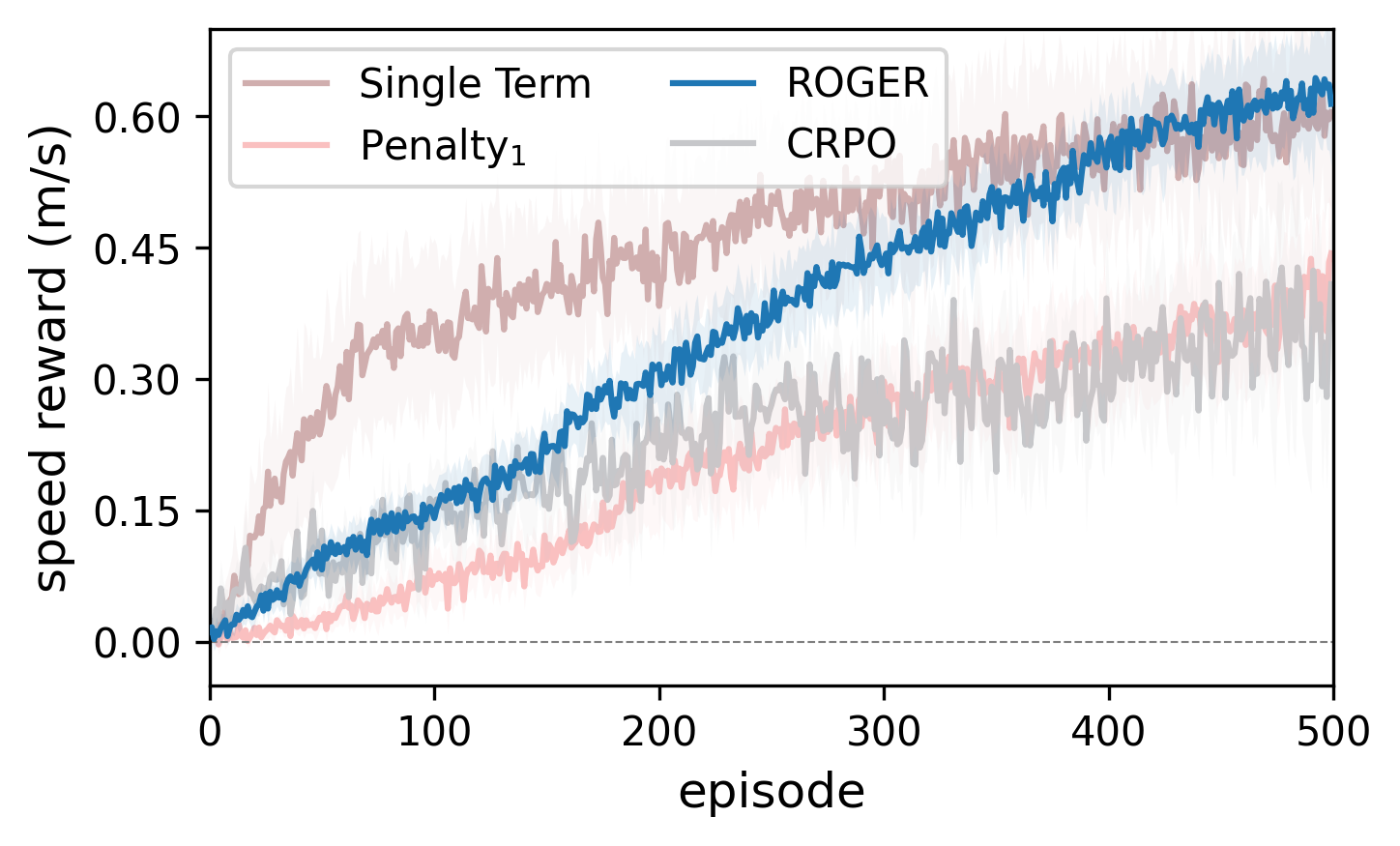}
	\caption{Evolution of the main reward term across 500 learning episodes from \revise{the} locomotion learning of the simulated Unitree B1 quadruped robot trained with (dark red) only the \main{} reward term, (light red) fixed gain penalty, (blue) \roger, and (gray) CRPO.}
	\label{fig:simquadrewardlines}
\end{figure}

\begin{figure*}[!t]
	\centering
	\includegraphics[width=\linewidth]{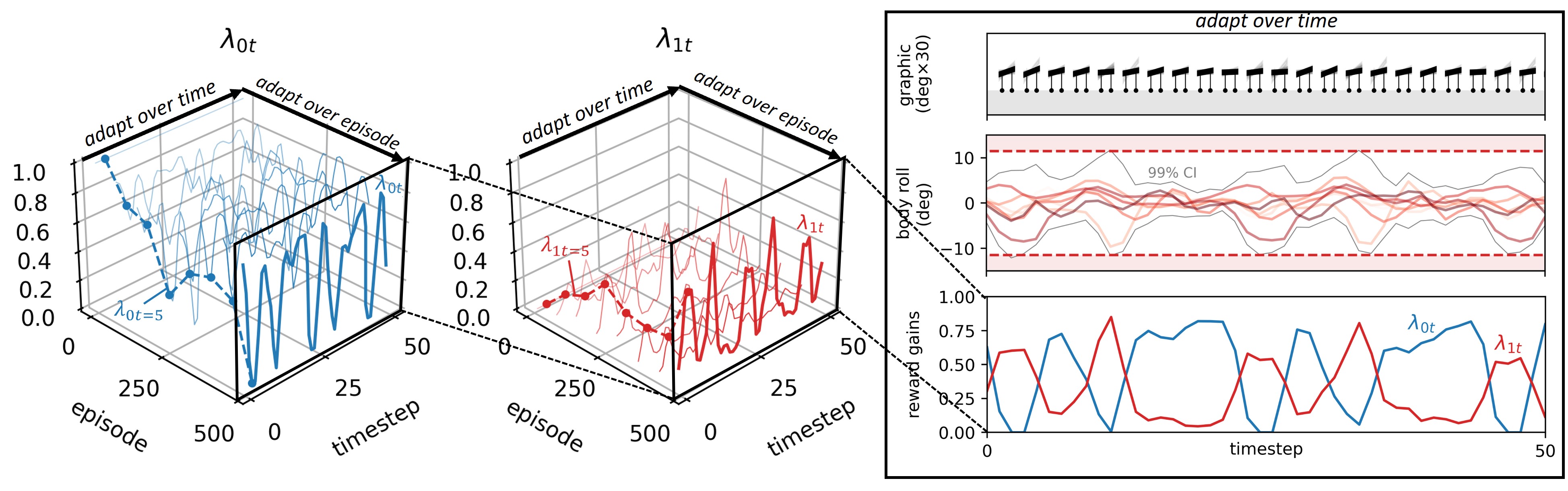}
	\caption{(Left) Evolution of the \main{} weighting gain (\var{\lambda}{0}) and body roll penalty weighting gain (\var{\lambda}{1}) across 50 timesteps and 500 episodes, obtained from the locomotion learning of a simulated quadruped robot. (Top right) Graphical illustration showing the robot body from the back view, with the values amplified 30 times for visualization purposes. (Middle right) Body roll \revise{values} collected over eight previous episodes (in red) \revise{along with} their 99\% confidence interval (in gray). (Bottom right) Evolution of the weighting gains: \var{\lambda}{0} in blue and \var{\lambda}{1} in red.}
	\label{fig:rewardadaptex}
\end{figure*}

Figures~\ref{fig:simquadroll} and \ref{fig:simquadpitch} also reveal that CRPO and \roger, which recompute weighting gains for each iteration, are the only two adaptive weighting approaches capable of \revise{satisfying constraints} throughout learning. CRPO had zero violations in 25,000 timesteps, while \roger{} merely exhibited one in 50,000 timesteps \revise{(without any falls; the violation was caused by exploration, as no violation was observed during reruns without exploration)}. This corresponds to an estimated violation probability of around $2\times10^{-12}$\%. In comparison, PDO exhibited $\approx$ 20\deg{} in orientation deviations and an estimated violation probability of 0.02, significantly higher than those from CRPO and \roger{} \pval{$\ll$ 0.01}{two-proportion test}. Interestingly, OL-AUX, designed to maximize the increase in the \main{} reward, exhibited similar 20\deg{} deviations with a lower estimated violation probability of 0.0006--1.6\% compared to PDO \pval{$\ll$ 0.01}{two-proportion test}. Nevertheless, the deviation and estimated violation probability of OL-AUX were still significantly greater than those of CRPO and \roger{} \pval{$\ll$ 0.01}{two-proportion test}.

  Among the three techniques with fewer than three violations in over 25,000 training timesteps presented in Figure~\ref{fig:simquadresult}, \roger{} achieved the highest final \main{} reward. The \main{} rewards reached 0.1 m/s and 0.3 m/s after 50 and 250 learning episodes, \revise{respectively,} before the final value of 0.6 m/s at 500 episodes, as shown in Figure~\ref{fig:simquadrewardlines}. This final value of 0.6 m/s was 50\% \revise{greater} than those of the fixed gain penalty and CRPO, which were $\approx$ 0.3 m/s \pval{$\ll$ 0.01}{t-test}. Interestingly, the final \main{} reward from \roger{} matched that \revise{achieved} when using the \main{} reward solely \pval{= 0.54}{t-test}, but with near-zero constraint violation. This highlights that \roger{} can autonomously and properly adapt and balance different reward weighting gains while ensuring constraint satisfaction.

 Finally, Figure~\ref{fig:rewardadaptex} presents the \revise{underlying} mechanism of \roger{} in more detail, revealing that the \main{} weighting gain \var{\lambda}{0} and roll penalty weighting gain \var{\lambda}{1} are dynamically adjusted based on the estimated body roll (gray lines), both across and within episodes. When the robot tilts to one side, \var{\lambda}{0} decreases and \var{\lambda}{1} increases to prevent tilting at that specific stage. Conversely, when the robot is stable and its body roll exhibits minor deviations, \var{\lambda}{0} increases and \var{\lambda}{1} decreases to prioritize the \main{} reward at that stage. Given the rhythmic nature of robot locomotion in this case, where the robot walks for two gait cycles in 50 timesteps, the weighting gains exhibit two periodic repetitions, offering insights into the learning objectives across different stages.

\subsubsection{Physical Robot Experiment} According to the results of the simulation experiment, \roger{} was then tested on a 60-kg physical quadruped robot. The robot was trained for 300 learning episodes with all mapping weights initialized as zero and the reward and penalty estimated from the outputs of an Intel RealSense T265 camera installed at the front. \revise{Training started from a standing posture---i.e., the home configuration---which has a high center of mass and is prone to falling.} A wheeled 15-kg support structure with slack ropes was employed as a safety mechanism. The support was \revise{set up} such that it did not assist in maintaining the stability of the robot during learning. However, it added complexity to the task, as the robot had to drag the structure while learning from an imperfect pose estimation. Due to the limited testing space, the robot waited for three seconds after each episode to allow the experimenter to pause the main program and reorient the robot after it went near the testing space boundaries. In total, the entire trial, including locomotion learning and reset, lasted almost one hour.



\begin{figure*}[t]
	\centering
	\begin{subfigure}[t]{0.29\linewidth}
		\centering
		\includegraphics[width=\linewidth]{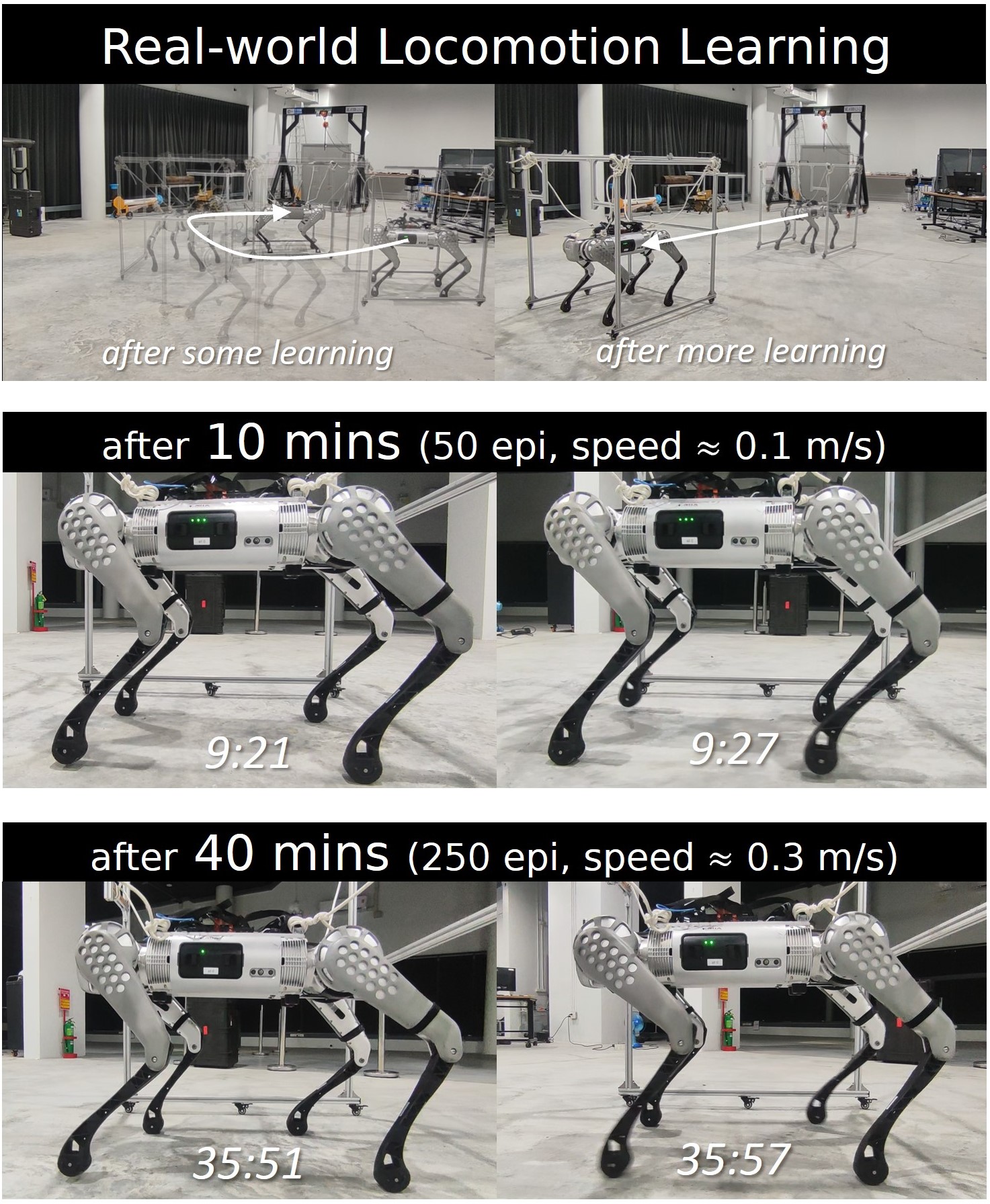}
		\caption{}
		\label{fig:realquadsnaps}
	\end{subfigure}
	\hfill
	\begin{subfigure}[t]{0.66\linewidth}
		\centering
		\includegraphics[width=\linewidth]{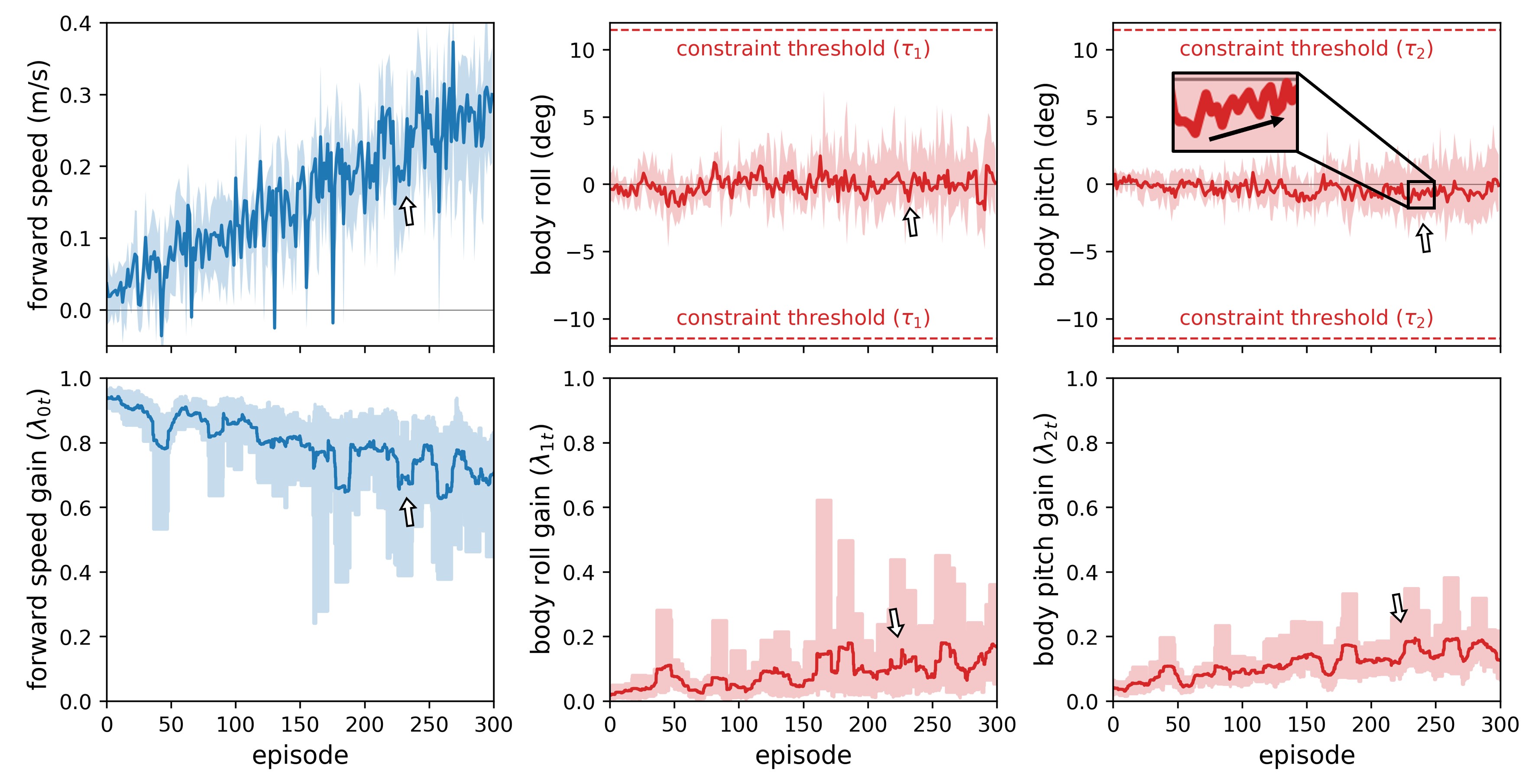}
		\caption{}
		\label{fig:realquadgraphs}
	\end{subfigure}
	\caption{(a) Snapshots capturing the locomotion learning of a physical robot after 50 and 250 episodes. (b, top) Forward speed (i.e., \main{} reward), robot roll, robot pitch, (b, bottom) the reward weighting gains, and their min-max ranges, recorded throughout 300 learning episodes. The gray solid line indicates zero forward speed, while the red dashed lines indicate the \cthres s. A video of this experiment is available at \realbonevideo.}
	\label{fig:realquadresult}
\end{figure*}

Figure~\ref{fig:realquadresult} presents the results of physical locomotion learning. The robot successfully completed all five trials without falling, demonstrating the capability of \roger{} to train a heavy quadruped robot in real-world conditions. By episode 50 ($\approx$ 10 minutes), the robot had achieved a forward speed of 0.1 m/s, with roll and pitch angles remaining below 3\deg. By episode 250 ($\approx$ 40 minutes), the average forward speed had increased to 0.3 m/s, matching the performance observed in simulation experiments and potentially suggesting a minimal sim-to-real learning performance gap. 

Notably, even under continuous random exploration, the robot effectively maintained its roll and pitch angles below the \cthres{} of 10\deg{} throughout 300 episodes, as shown in the top row of Figure~\ref{fig:realquadgraphs}. This stability was achieved through the dynamic adjustment of reward weighting gains. In general, the orientation deviation increased slightly as the robot learned, reducing the \main{} weighting gain and increasing the roll and pitch penalty weighting gains, as shown in the bottom row of Figure~\ref{fig:realquadgraphs}. \revise{As shown in the video (\realbonevideo), the robot learned relatively short steps with low foot lift while walking on flat terrain, a strategy that helps maintain stability under continuous exploration. Longer steps and higher foot lifts, while potentially more efficient, are more likely to violate stability constraints during such stochastic movements, as shown in \simbonevideo. Nevertheless, Figure~\ref{fig:intro} illustrates that a variety of complex gaits emerged under more challenging conditions, supporting that physical interaction with the environment shapes the resulting behavior \cite{howbodyshapethewaywethink}.}

Apart from the long-term trend, specific adjustment\revise{s} in the weighting gains \revise{were} also observed throughout the learning in Figure~\ref{fig:realquadgraphs}. For example, an interesting event occurred around episode 220 ($\approx$ 2:36 minutes in the video; \realbonevideo), when the robot started oscillating in both roll and pitch angles. In response, \roger{} lowered the \main{} weighting gain and increased the penalty weighting gains, as shown in the bottom row of Figure~\ref{fig:realquadgraphs}. The robot then spent the next five episodes adjusting its locomotion pattern to reduce the oscillation. This is reflected in a slight decrease in forward speed, accompanied by roll and pitch values changing toward zero shortly after 220 episodes. After 3:00 minutes, the robot had successfully reached a stable gait, demonstrating the autonomous balancing process of the \main{} reward and constraint penalty terms in real time.

\revise{To further investigate this mechanism, the robot was trained under four challenging conditions, as shown in Figure~\ref{fig:intro}(b--e): (1) a step field with step heights varying between 4 and 10 cm; (2) a gravel field with gravel diameters ranging from 2 to 7 cm; (3) a slippery surface, created by applying machine lubricant to a whiteboard (static friction coefficient $\approx$ 0.25); and (4) dynamic loading, involving a 1.5-kg bottle of water placed at the front, two 1.5-kg bottles at the rear, and a football sack---together comprising nearly 10\% of the robot's weight.}

\revise{Under dynamic loading, the robot successfully maintained balance while walking forward. On the slippery surface, initial difficulty in maintaining ground contact was observed, but the robot eventually learned to slide forward while preserving body posture. On the loose gravel terrain, the robot leaned slightly forward and learned to kick the top-layer gravel to clear a path. In the step field, the terrain's rigidity restricted both foot placement and terrain manipulation, initially leading to unsuccessful attempts. To deal with this, training was initialized with a trotting-in-place gait featuring increased foot lifting. As a result, the robot developed a hopping-like gait to overcome the steps. Notably, all these behaviors emerged without the use of any exteroceptive terrain sensing, suggesting that feasible control policies should be reachable within the exploration while learning should satisfy constraints throughout.}

\subsection{Hopper Locomotion Learning}
\label{sec:hopexp}

This experiment extends the evaluation of \roger{} to further investigate a condition of assumption violations. A simulated MuJoCo hopper was used, which requires a highly dynamic hopping gait and is more sensitive than a quadruped robot. Due to being feedback-dependent, which cannot currently be achieved with the SME architecture \cite{smeagol}, the hopper was instead controlled by a three-hidden-layer neural network with 256 hyperbolic tangent neurons per layer, trained using a standard PPO algorithm \cite{ppo}, as shown in Figure~\ref{fig:quaddiagram}. The experiment followed the default OpenAI Gymnasium setup, except that the reward was decomposed into multiple channels as required by \roger{}. The reward/penalty terms included forward velocity plus a healthy reward as the \main{} reward, an absolute torque penalty to penalize torque usage, and an absolute body orientation penalty to encourage an upright body. The torque penalty represents a non-zero constraint in the optimal solution (i.e., zero-penalty optimality violated), while the orientation penalty represents a sensitive constraint (i.e., gentle system and transition dynamics violated). The penalty thresholds were set at $\tau_1 = 1.0$ (maximum value) for torque and $\tau_2 \approx 10^\circ$ for orientation. Other hyperparameters, e.g., those of the neural network and learning algorithm, were empirically selected such that the robot could complete the task with the default reward function.

\revise{Three testing conditions were evaluated: (1) the baseline reward functions with fixed weighting gains obtained directly from the OpenAI Gymnasium, i.e., default; (2) CRPO; and (3) \roger{}.} Each condition was repeated for over 20 trials, with each trial lasting one million timesteps and 2048 timesteps per episode. Four performance metrics were employed: hopping distance (i.e., related to the main reward), absolute joint torque, orientation deviation, \revise{and the percentages of constraint violations.}

\begin{figure}[!h]
	\centering
	\includegraphics[width=0.9\linewidth]{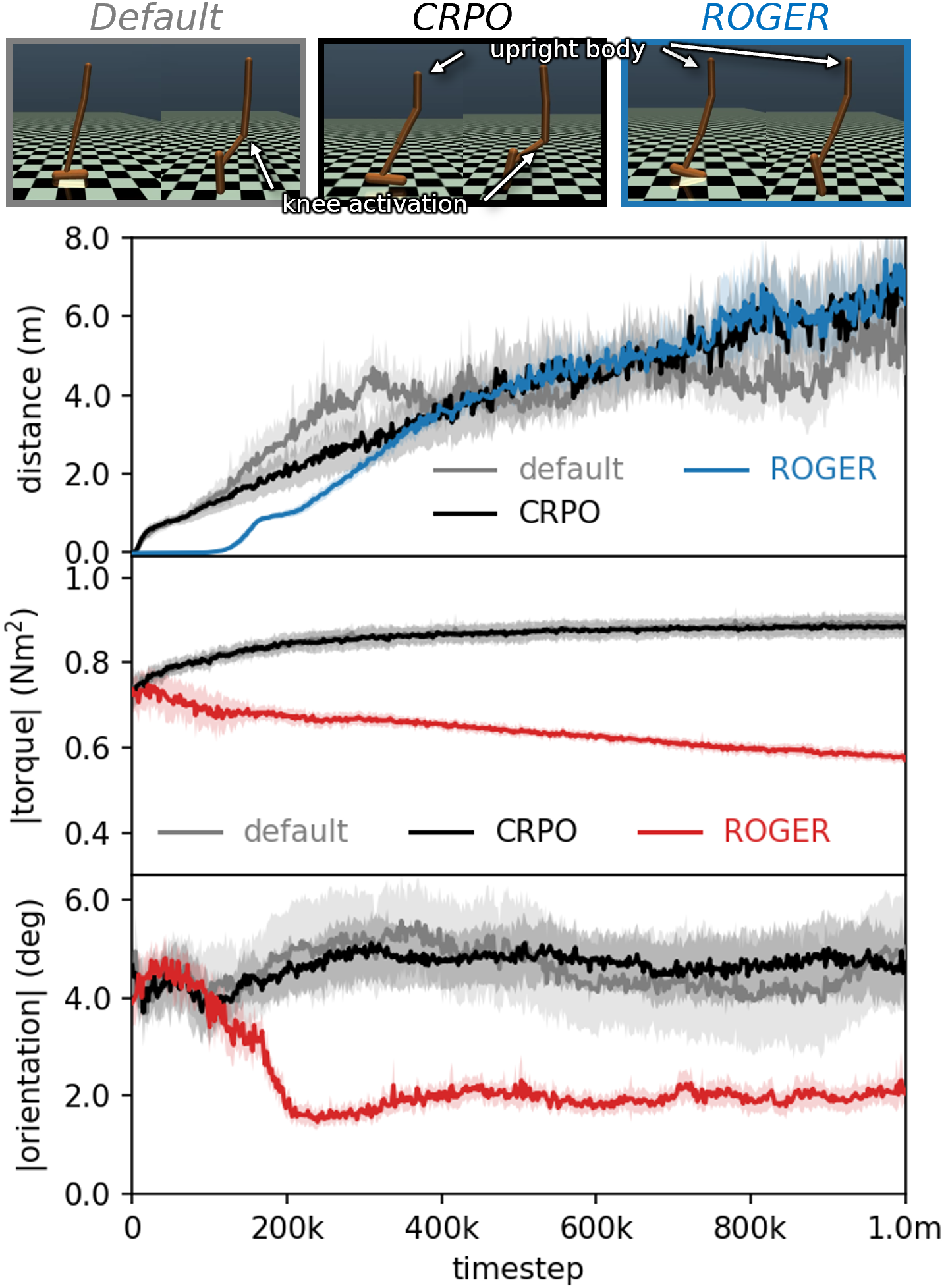}
	\caption{\revise{Snapshots, hopping distance, joint torque usage, and absolute orientation deviation, obtained from a MuJoCo hopper trained using the default reward function from OpenAI Gymnasium, CRPO \cite{crpo}, and \roger.} A video of the experiment is available at \hoppervideo.}
	\label{fig:hoppercompareresult}
\end{figure}

Figure~\ref{fig:hoppercompareresult} demonstrates that overall, \roger{} significantly outperformed the baselines, achieving \revise{greater} hopping distances, reduced torque usage, and smaller orientation deviations. Specifically, \roger{} achieved a hopping distance of 6.6~m, representing a 20\% improvement over the default reward function \pval{$< 0.01$}{t-test}, \revise{which was similar to that of CRPO.} With \roger, the average torque usage was 0.33~Nm, representing a 57\% reduction from the default reward function \pval{$< 0.01$}{t-test} \revise{and a 58\% reduction from CRPO \pval{$< 0.01$}{t-test}}. Similarly, the average orientation deviation was 2\deg, representing a 59\% decrease compared to the default reward function \pval{$< 0.01$}{t-test} \revise{and 53\% decrease compared to CRPO. In total, the percentages of constraint violation obtained with \roger{} were lower than those with the default reward function and CRPO both throughout the learning process and in the last 10 episodes, as summarized in Appendix.} However, unlike the quadruped experiment, while \roger{} ensures that the expected value of constraints is satisfied, the absence of falls was not guaranteed due to assumption violations (i.e., having sensitive system dynamics). Nevertheless, it offers improved performance and a simpler tuning process.

Figure~\ref{fig:hoppercompareresult} further reveals the relationship between the three reward/penalty channels and the underlying mechanism of \roger. During the first 50,000 timesteps, the hopper focused on reducing torque usage and orientation penalties, as indicated by their steady decline. Once these terms had been sufficiently minimized, the hopper shifted its focus toward maximizing the \main{} reward, resulting in a slight delay in \main{} reward maximization. \revise{Subsequently}, the hopper showed a continuous performance increase while maintaining low torque usage and minimal orientation deviation.

Figure in Appendix suggests that, \revise{since} \roger{} utilizes embodied interaction between itself and the environment to adapt reward weighting gains, the results may be environment-specific. The torque usage obtained from \roger$_\tau$, i.e., with only the \main{} reward and torque penalty, mirrored that obtained from the three-channel \roger, with the orientation deviation being similar to that obtained from the default reward function. Conversely,  the orientation deviation obtained from \roger$_\beta$, i.e., with only the \main{} reward and orientation penalty, matched that of the three-channel \roger, with the torque usage being similar to that obtained from the default reward function.

Interestingly, \roger{} also demonstrated its success under stricter constraints, with the \cthres s being halved to $(\tau_1,\tau_2) = (0.5,5^\circ)$, as shown in Appendix. In this scenario, the hopper initially prioritized the reduction of constraint penalties before maximizing the \main{} reward. Although this stricter configuration delayed reward optimization, the hopper maintained forward hopping without moving backward, i.e., exhibiting a negative hopping distance. These results reveal that given sufficient learning, \roger{} can achieve comparable performance to finely tuned reward functions with constraint satisfaction.

\section{Conclusion} 
\label{sec:conclusion}

Nowadays, robot locomotion learning is not yet commonplace in the real world due to certain requirements, such as avoiding falling while learning in the case of quadruped robots \cite{walkinthepark}. Current state-of-the-art RL techniques often experience constraint violations in quadruped robots, especially during learning \cite{trustqp_gofeasible,iros_ethcompare}, while selecting proper reward weighting gains is time-consuming \cite{hyperandhowtotunethem} and risky, as shown in Appendix. Therefore, this work introduces \rogerf{} (\roger), a simple rule that adjusts reward weighting gains and balances different objectives through the dynamic interaction between the robot and the environment. \roger{} increases the penalty gains and decreases the \main{} weighting gain as the robot approaches the \cthres s, continuously enforcing constraint satisfaction. Conversely, it increasingly relaxes the constraints to prioritize \main{} reward maximization as the robot moves away from the \cthres s. Unlike most state-of-the-art techniques, \revise{which heavily rely on simulation-based training \cite{cbf,dd,cpo,crpo}, \roger{} strictly enforces constraints, which potentially slows down learning in extreme conditions (Appendix) but is crucial for real-world learning where violations are catastrophic. Moreover,} \roger{} allows intuitive hyperparameter selection: \cthres s ($\tau_i$; acceptable values) and uncertainty levels (\revise{$k_\sigma$, e.g., $k_\sigma = 3$} for 99.9\% confidence), \revise{making the setting of unrealistic constraints, e.g., 1$^\circ$ body deviation, unlikely in practice.}

\roger{} demonstrates its effectiveness in both theory and practice. Theoretically, constraint satisfaction is achieved through partial stability under key conditions: near the \cthres s and when converged, while ensuring an increase in the \main{} reward, as detailed in Appendix. \revise{Although \roger{} is designed around three key assumptions---zero-penalty optimality, gentle system dynamics, and gentle learning dynamics---it still performs well even when these assumptions are violated. First, despite the torque constraint violation, where optimal torque values are non-zero for locomotion, \roger{} remains effective, as shown in Figure~\ref{fig:hoppercompareresult}. Second, in tests with the MuJoCo hopper and walker2D, which are less stable than the quadruped robot, \roger{} outperforms state-of-the-art methods under sensitive system dynamics, although fall prevention isn't guaranteed (see Appendix). Finally, an experiment with excessive learning rates on a quadruped robot, which has gentle system dynamics compared to the hopper and walker2D, reveals that sensitive learning dynamics can lead to failures, as constraint satisfaction cannot be guaranteed (Appendix)}

Experimentally, \roger{} exhibited near-zero violation throughout the multiple-seed locomotion learning of a heavy quadruped robot (i.e., five times heavier than Unitree A1 \cite{walkinthepark}), \revise{which is comparable to carefully tuned fixed weighting gains and state-of-the-art CRPO \cite{crpo}, but with a higher performance}. Furthermore, the technique can be applied to DNN-PPO \cite{anymalparkour}, outperforming the default reward functions in MuJoCo continuous locomotion learning tasks both in terms of the \main{} reward (e.g., distance), constraints where optimal values are expected to be non-zero (e.g., torque usage), and constraints where optimal values are expected to be \revise{near} zero (e.g., orientation stability). Finally, \roger{} offers stable real-world locomotion learning on a quadruped robot, \revise{both on regular terrain and under various challenging conditions,} in less than an hour, with the robot neither falling nor requiring extensive tuning, \revise{even with limited training data (i.e., $\approx$ 500 timesteps per update, or rather 0.5\% of that in \cite{massiveparallel}) and without exteroceptive terrain sensing.}

In summary, this work highlights how embodied interaction can dynamically adapt reward weighting gains in real time, achieving near-optimal performance across all tests presented here while minimizing the need for reward weighting gain tuning. By enabling continuous improvement and simplifying tuning, this work contributes to physical AI research and development in terms of efficient continual robot learning in the real world, especially when parameter selection is high-stake\revise{s} or resource-intensive. \revise{Thus, if constraint violations are acceptable during learning, simulation-based methods may be preferable; for real-world continual learning or fine-tuning, \roger{} may be a better choice.}


\section{Limitation}
\label{sec:limit}
Currently, \roger{} is limited to proprioceptive constraints within robot systems that satisfy zero-penalty optimality as well as gentle system and learning dynamics. It does not account for exteroceptive constraints like global positioning, which can accumulate over time. These limitations, including real-world quadruped locomotion learning on complex terrain with exteroceptive terrain sensing \cite{gollum}, will be addressed in future works.

\section{Appendix}
Appendix is available at: \newline \href{https://doi.org/10.15607/RSS.2025.XXI.123}{https://doi.org/10.15607/RSS.2025.XXI.123}

\bibliographystyle{unsrtnat}



\end{document}